\documentclass[10pt,letterpaper]{article}
\usepackage[top=0.85in,left=1.25in,footskip=0.75in,marginparwidth=2in]{geometry}

\usepackage[utf8]{inputenc}
\usepackage[english]{babel}
\usepackage[flushleft]{threeparttable} 
\usepackage{amsmath}
\usepackage{algorithm} 
\usepackage{algorithmic} 
\usepackage{makecell} 
\usepackage[toc, page]{appendix}

\usepackage{cite}

\usepackage{nameref,hyperref}

\usepackage[right]{lineno}

\usepackage{microtype}
\DisableLigatures[f]{encoding = *, family = * }

\raggedright
\usepackage{changepage}

\usepackage[aboveskip=1pt,labelfont=bf,labelsep=period,singlelinecheck=off]{caption}

\makeatletter
\renewcommand{\@biblabel}[1]{\quad#1.}
\makeatother

\usepackage{lastpage,fancyhdr,graphicx}
\usepackage{epstopdf}
\fancyheadoffset[L]{2.25in}
\fancyfootoffset[L]{2.25in}

\usepackage{color}

\definecolor{Gray}{gray}{.25}

\usepackage{graphicx}

\usepackage{sidecap}

\usepackage{wrapfig}
\usepackage[pscoord]{eso-pic}
\usepackage[fulladjust]{marginnote}
\reversemarginpar

\begin{document}
\vspace*{0.35in}

\begin{flushleft}
{\Large
\textbf\newline{On the Statistical Challenges of Echo State Networks and Some Potential Remedies}
}
\newline
\\
Qiuyi Wu\textsuperscript{1,$\star$},
Ernest Fokoue\textsuperscript{1,$\dagger$},
Dhireesha Kudithipudi\textsuperscript{1,$\ddagger$},
\\
\bigskip
\bf{1} Rochester Institute of Technology, USA
\\
\bigskip
$\star$ qw9477@rit.edu\\
$\dagger$ ernest.fokoue@rit.edu\\
$\ddagger$ dxkeec@rit.edu

\end{flushleft}

\section*{Abstract}
Echo state networks are powerful recurrent neural networks. However, they are often unstable and shaky, making the process of finding an good ESN for a specific dataset quite hard. Obtaining a superb accuracy by using the Echo State Network is a challenging task. We create, develop and implement a family of predictably optimal robust and stable ensemble of Echo State Networks via regularizing the
training and perturbing the input. Furthermore, several distributions of weights have been tried based on the shape to see if the shape of the distribution has the
impact for reducing the error.
We found ESN can track in short term for most dataset, but it collapses in the long run. Short-term tracking with large size reservoir enables ESN to perform strikingly with superior prediction. Based on this scenario, we go a further step to aggregate many of ESNs into an ensemble to lower the variance and stabilize the system by stochastic replications and bootstrapping of input data.

\section*{Introduction}
As a class of Reservoir Computing model, Echo State Networks accurately track and predict time series data by Machine Learning techniques. The realizations of
the reservoirs in real-world applications are various, such as object recognition, speech recognition, robotic movement control, dynamic pattern classification, and chaotic time-series generation etc. The main challenge ESN faces now is the stability. ESN is notoriously difficult to
track data because of the following reasons: the performance of the base learner is quite unsettling, which demands tremendous efforts for tweaking the parameter in the initial setting stage; the system would collapse when the dataset contains much noise; ESN can only track temporarily and it unable to forecast in the long run.

Since Echo State Networks first proposed by Jager (2007)\cite{jaeger_2007}, many modified models have been created to enhance the performance of the original base model in recent years. Kudithipudi et al.(2015)\cite{kudithipudi2015design} designed a toroidal ESN architecture with hybrid topology to lower power consumption. Ma et al. (2016)\cite{ma_shen_chen_wang_wei_yu_2016} proposed Functional Echo State Networks (FESN) to enhance the separability of different classes in a high-dimensional functional space. Mcdermott and Wikle (2017)\cite{mcdermott_wikle_2017} constructed Quadratic Echo State Networks for ensemble building so that the machine can handle uncertainty in nonlinear spatio-temporal forecasting. 

The main contribution of this paper is to state the challenges of Echo State Networks and provide potential remedies to improve the performance of the system by building ensemble of ESNs for robustly stable and highly predictive machine. We create, develop and implement a family of predictably optimal robust and stable
ensemble of Echo State Networks via regularizing the training and perturbing the input. Furthermore, constructing several distributions of weights based on the
shape to see if the shape of the distribution has the impact for reducing the error. ESN can track in short term for most dataset, but it collapses in the long run. Short-term tracking with large size reservoir enables ESN to perform strikingly with superior prediction. Based on this scenario, we go a further step to aggregate many of ESNs into an ensemble to lower the variance and stabilize the system. We create the ensemble of Echo State Networks with two scenarios: (i) Adaptation on structured pattern matrices (ii) Stochastic replications and bootstrapping of input data.

The remainder of this paper is organized as follows. Section 2 gives brief introduction of Echo State Networks and then discusses the challenges we encounter to obtain superb accuracy in prediction from mainly three aspects: randomness in weights; distribution of weights; environmental and parameter setting. Section 3 and 4 elaborate the challenges demonstrated in the previous chapter and provide experimental results based on different datasets. Section 3 focus on the exploration of weight distribution. Section 4 investigates environmental setting the for optimal performance. Section 5 illustrates the tremendous improvement in prediction achieved by Ensembles of Echo State Network. Section 6 provides some final conclusions and directions for future work.

\section{Echo State Networks}
Traditional Neural Networks provide a large family of function approximators able to handling the situations where the underlying true function is heavily nonlinear. It is done by statistically modeling in a way that attempts to mimic the human brain. Recurrent Neural Networks go further step to allow the system handle time dependent data due to the distributed hidden state and non-linear dynamics. The distributed hidden state can efficiently store large past information and the non-linear dynamics are able to update the hidden state in sophisticated ways. \cite{sutskever_hinton_2010}

Proposed by Jager (2007)\cite{jaeger_2007}, Echo State Networks as one of Recurrent Neural Networks (RNNs) hold the idea of merely learning the output layer with early layers random and fixed. Typical Echo State Networks embody three layers: an input layer, a hidden layer and an output layer. Consider a typical ESN with K, N, L number of units in its input layer, hidden layer and output layer accordingly. The input units at time step n is denoted as $\mathbf{u}(n) = [u_1(n),u_2(n),...,u_K(n)]$, the units in the hidden layer is $\mathbf{x}(n) = [x_1(n),x_2(n),...,x_N(n)]$, and $\mathbf{y}(n) = [y_1(n),y_2(n),...,y_L(n)]$ is the units in the output layer. The input weights $\mathbf{W}^{in}$ with the form of $N\times K$ matrix connect the input units and the hidden units, and the internal weights $\mathbf{W}$ with the form of $N\times N$ matrix connect the internal units,
both are randomly generated. Only the output weights $\mathbf{W}^{out}$ with the form of $L\times (K+N+L)$ matrix connect the internal units and the output units are trained. The Figure 1 below showed all possible connections of Echo State Networks, including the feedback from $y(n-1)$ to $\hat{x}(n)$, the connection straight from the input to the output, and also the connection between output units.
\begin{figure}[!htbp]
\begin{center}
\includegraphics[scale=0.5]{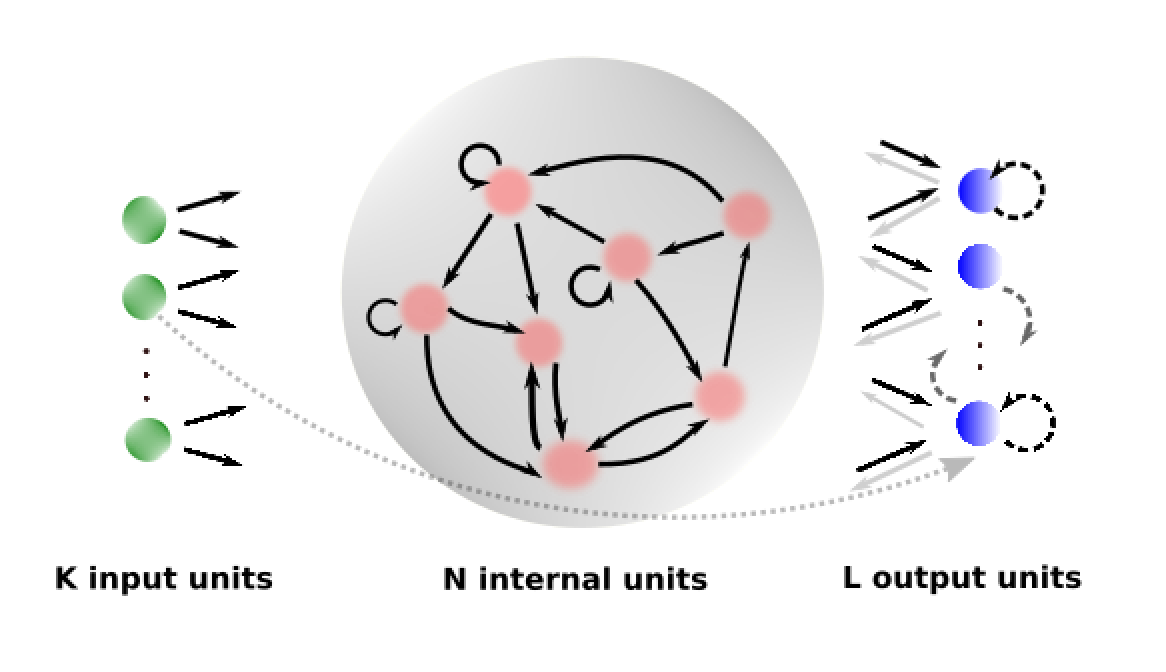} 
\end{center}
\caption{Typical Echo State Networks With All Possible Connections}
\end{figure}
For training typical ESNs, two steps are contained: first is to have the states updated from the previous step, second is to train the output weight $\mathbf{W}^{out}$:
\begin{align}
\mathbf{x}(n) &= \mathbf{f}(\mathbf{W}^{in}\mathbf{u}(n)+\mathbf{W}\mathbf{x}(n-1) + \mathbf{W}^{back}\mathbf{y}(n-1))\\
\mathbf{y}(n) &= \mathbf{f}^{out}(\mathbf{W}^{out}(\mathbf{u}(n), \mathbf{x}(n), \mathbf{y}(n-1)  ))
\end{align}
where $\mathbf{f} = (f_1,f_2,...,f_N)$ are the activation functions of hidden units (usually sigmoid function, Hyperbolic Tangent or ReLU function in ESN) and $\mathbf{f}^{out} = (f^{out}_1, f^{out}_2,..., f^{out}_L)$ are the activation function of output units (usually linear function). Only the output layer weights need to be trained with simple linear regression algorithms or others learning algorithm. Thus echo state network models have simplified training algorithms compared to other recurrent neural networks.

\section{Challenges of ESN}
Traditional recurrent neural network can process a sequence of inputs dependent on each other but may encounter the issue of vanishing gradient problem and these algorithms are suffering from memory storage requirement training time restriction. To remedy these problems, echo state network can be used as a partially-trained recurrent neural network. It can handle the issues by having a relatively large reservoir with sparsely connected neurons. The connections in the reservoir are randomly assigned and the weights in the input layer and the reservoir layer are not trained. The advantage of this approach is in the incredibly simple training procedure since most of the weights are assigned only once and at random. However, the simplicity in training also bring challenges in robustness. Randomness assigned in the reservoir creates instability and yields high variance. In addition, the base learner demands complicated tuning in parameter setting, which requires tremendous efforts. Some questions come up along with challenges:
What kind of randomness we can try? Does uniform random weight or Gaussian random weight matter? What’s the range of random number? Shall we use fixed leaking rate or dynamic one? In this section we mainly explore two aspects of challenges: (I) Randomness in weights (II) Parameter turning efforts for optimal environment.

\subsection{Randomness in Weights}
The randomness in weights brings potential instability and creates high variance in the machines. Echo State Networks (ESN) are particularly sought-after because of the inherent structure not needing heavily computational back propagation. Traditional Neural Networks use back propagation with expensive computation process, while Echo State Networks avoid this painful step.  Unfortunately, ESN replaces back propagation with random assignments or values to reservoir oscillators. 
\begin{figure}[H]
\begin{center}
\includegraphics[scale=0.25]{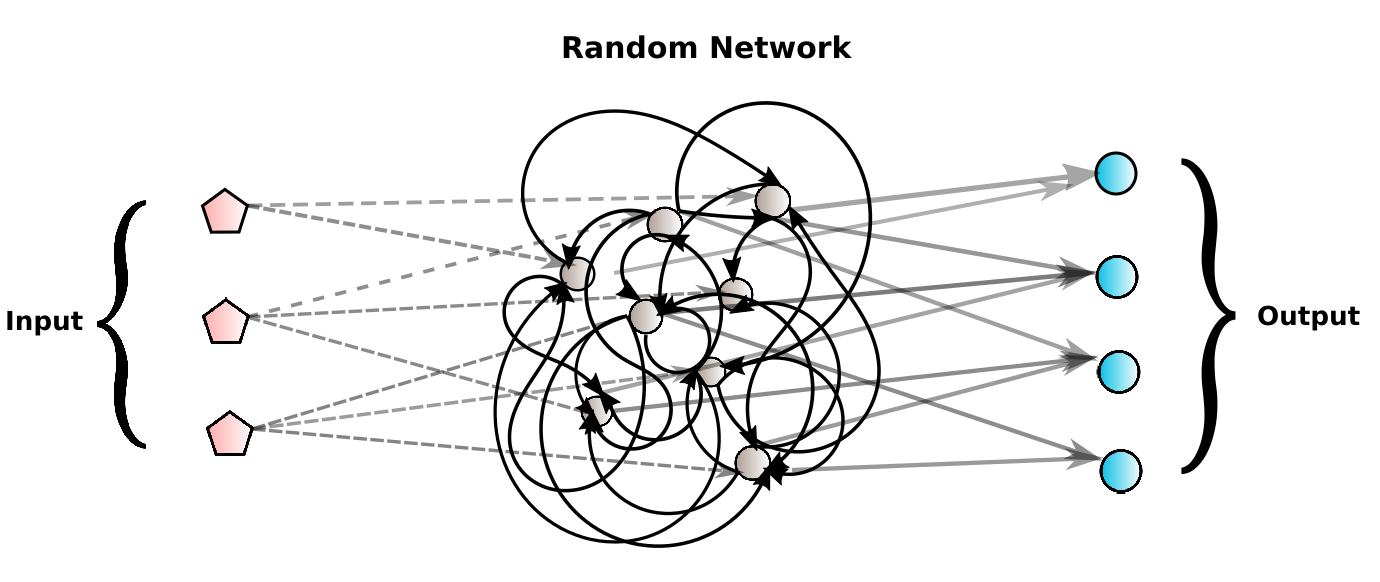} 
\end{center}
\caption{Random Assignment In Simple ESN}
\end{figure}

What sort of distribution to be used in random assignments? To explore the randomness of the weights, several distributions are worthy being tried. Here based on the shape we explored four kinds of distributions: Uniform distribution, Gaussian distribution with the same variance as the uniform distribution, Gaussian distribution with the same range as the uniform distribution, Arcsine distribution. All of them have significant difference, it is interesting to see if the shape of the distribution has the impact for reducing the error.
\begin{figure}[H]
\begin{center}
\includegraphics[scale=0.25]{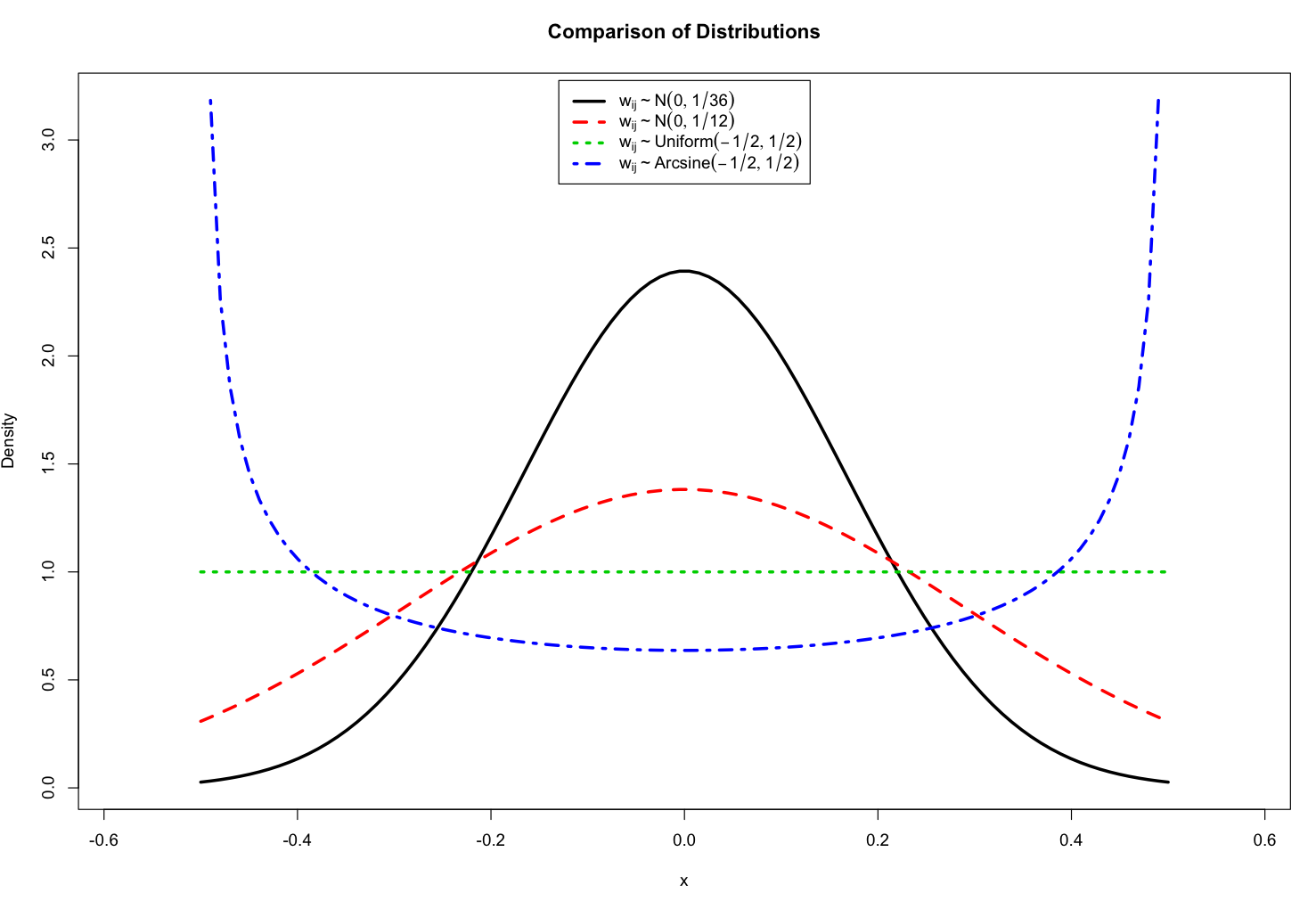} 
\end{center}
\caption{Probability Density Function of Four Different Distributions}
\end{figure}

\subsection{Preprocessing for Proper Scaling}
Similar to other neural networks in environment setting, there are too many parameters and hyper-parameters to be tuned and adjusted. There are several ways for data preprocessing and parameter tuning: 
\begin{itemize}
\item Data preprocessing\\
Several ways can be used to scale the dataset as preprocessing methods. As far as preprocessing on the input space is concerned, wherever deemed necessary, we may consider the following for each entry $x_{ij}$  of $\mathbf{X}$:
\begin{itemize}
\item Cubitize:
$$ x_{ij} \leftarrow \frac{x_{ij}-min(\mathbf{x}_j)}{max(\mathbf{x}_j)-min(\mathbf{x}_j)} $$
\item Standardize: 
$$ x_{ij} \leftarrow \frac{x_{ij}-mean(\mathbf{x}_j)}{\sqrt{variance(\mathbf{x}_j)}} $$
\item Unitize:: 
$$ x_{ij} \leftarrow \frac{x_{ij}-mean(\mathbf{x}_j)}{\|\mathbf{x}_j-mean(\mathbf{x}_j)\mathbf{l}_p \|_2} $$
$$\textup{where }\|x_j-mean(\mathbf{x}_j)\mathbf{l}_p\|_2 = \sqrt{\sum_{i=1}^n[x_{ij}-mean(\mathbf{x}_j)]^2} $$
\end{itemize}
\item Dynamic leaking rate\\ 
Leaking rate $\alpha$ in Echo State Networks in the rang of $(0,1]$ is considered as the speed of reservoir update dynamics in continuous time \cite{lukosevicius_2012}. The choice for leaking rate effects the performance dramatically. In addition, the status of leaking rates as fixed or dynamic can also bring different results. The update process is:
\begin{align}
\tilde{x}[n] &= tanh(W_{in}u[n]+Wx[n-1])\\
x[n]&=(1-\alpha)x[n-1] + \alpha\tilde{x}[n]\\
y[n] &= f^{out}(W_{out}(u[n], x[n]))
\end{align}
when $\alpha = 1$, it's the special case as $x[n]=\tilde{x}[n]$ the model does not have leaky integration.
\item Perturbation on each time step\\
The idea of adding random perturbation, also known as white noise, to the the input at each time step comes from perturbation theory, a general mathematical methods. We add a small random perturbation to the input as they will become the initial conditions for the next time step integration. After adding perturbation $\tau$ the update process is:
\begin{align}
\tilde{x}[n] &= tanh(W_{in}u[n]+Wx[n-1] + \tau)\\
x[n]&=(1-\alpha)x[n-1] + \alpha\tilde{x}[n]\\
y[n] &= f^{out}(W_{out}(u[n], x[n]))
\end{align}
\item Perturbation on initialization\\
In the initial step, the value of $x[0]$ is set as 0 by default. But now we give it small perturbation with different distributions to see if the small variability can make any difference:
\begin{verbatim}
  x0 = rep(0,resSize)           # original value
  x0 = rnorm(resSize, 0, 0.1)   # perturbation scenario 1
  x0 = runif(resSize,0,1)       # perturbation scenario 2
\end{verbatim}
So the update process is:
\begin{align}
x[0] &= x_0,\qquad x_0 \neq 0\\
\tilde{x}[n] &= tanh(W_{in}u[n]+Wx[n-1])\\
x[n]&=(1-\alpha)x[n-1] + \alpha\tilde{x}[n]\\
y[n] &= f^{out}(W_{out}(u[n], x[n]))
\end{align}
\end{itemize}

\section{Effect of Weight Distribution}
As mentioned in the previous chapter, we have explored four different kinds of distributions of weights based on shapes.
The most common distribution many paper used is Uniform distribution. Gaussian distributed weights is another popular ones used in Echo State Network. Here we take two elements into account for the comparison with the Uniform distributed weights. We try to control the range of the distribution or the variance of the distribution, to see the which kind of Gaussian distributed weights have better predictive capability. Arcsine distribution is noticeable for its U shape. Based on the original dataset from time-series data Mackey-Glass (delay=17) by Mantas Lukosevicius \cite{lukosevicius_2012}, we explore four different scenarios.
Dataset Mackey-Glass time series refers to the following delayed differential equation:
\begin{align}
\frac{dx}{dt}=\beta x(t)+\frac{\alpha x(t-\tau)}{1+x(x-\tau)^{10}}
\end{align}
where x is the series in the time t, and $\tau$ is the time delay. In this case time delay is 17.
\begin{figure}[H]
  \begin{center}
  \includegraphics[width=4in]{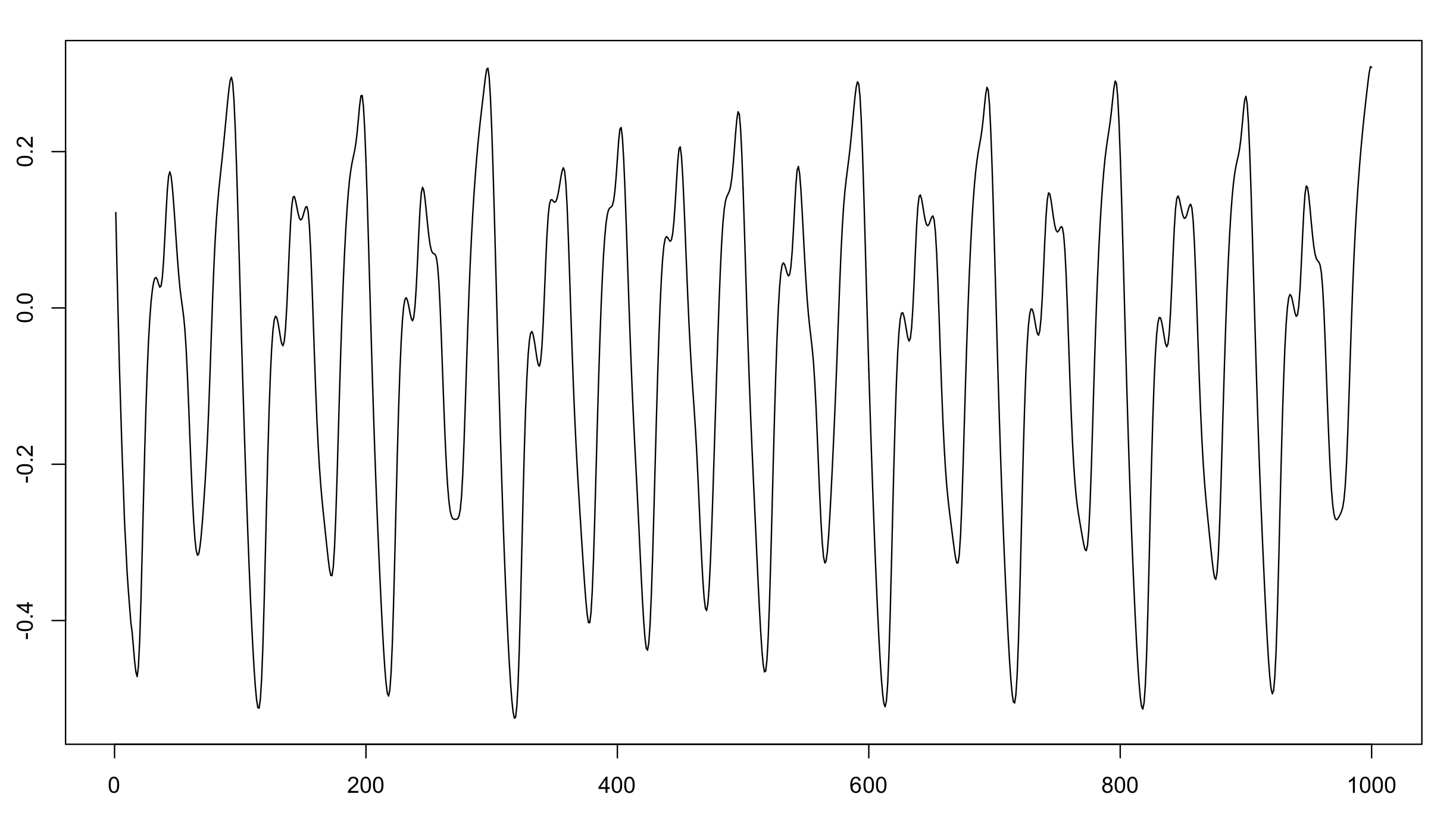}
  \caption{Sample of Mackey-Glass (delay=17) Data}
  \end{center}
\end{figure}

\subsection{Uniform Distributed Weights}
The initial attempt for the option of distributed weights is uniform distribution due to its simple and interpretable form. Typically the error measure used here is mean-absolute-error (MAE) or mean-square-error (MSE) or root-mean-square-error (RMSE) showed below, and we use MSE for in this paper.
\begin{align}
E_{MAE}(\mathbf{y},\mathbf{y}^{target}) &= \frac{1}{T}\sum_{i=1}^T|y_i-y_i^{target}|\\
E_{MSE}(\mathbf{y},\mathbf{y}^{target}) &= \frac{1}{T}\sum_{i=1}^T(y_i-y_i^{target})^2\\
E_{RMSE}(\mathbf{y},\mathbf{y}^{target}) &= \sqrt{\frac{1}{T}\sum_{i=1}^T(y_i-y_i^{target})^2}
\end{align}
So here for Uniform distributed weights (in both input layer and hidden layer). From the plot below shows the predictive capability of the machine for the target signal performs very well. 
\begin{figure}[H]
  \begin{center}
  \includegraphics[width=4in]{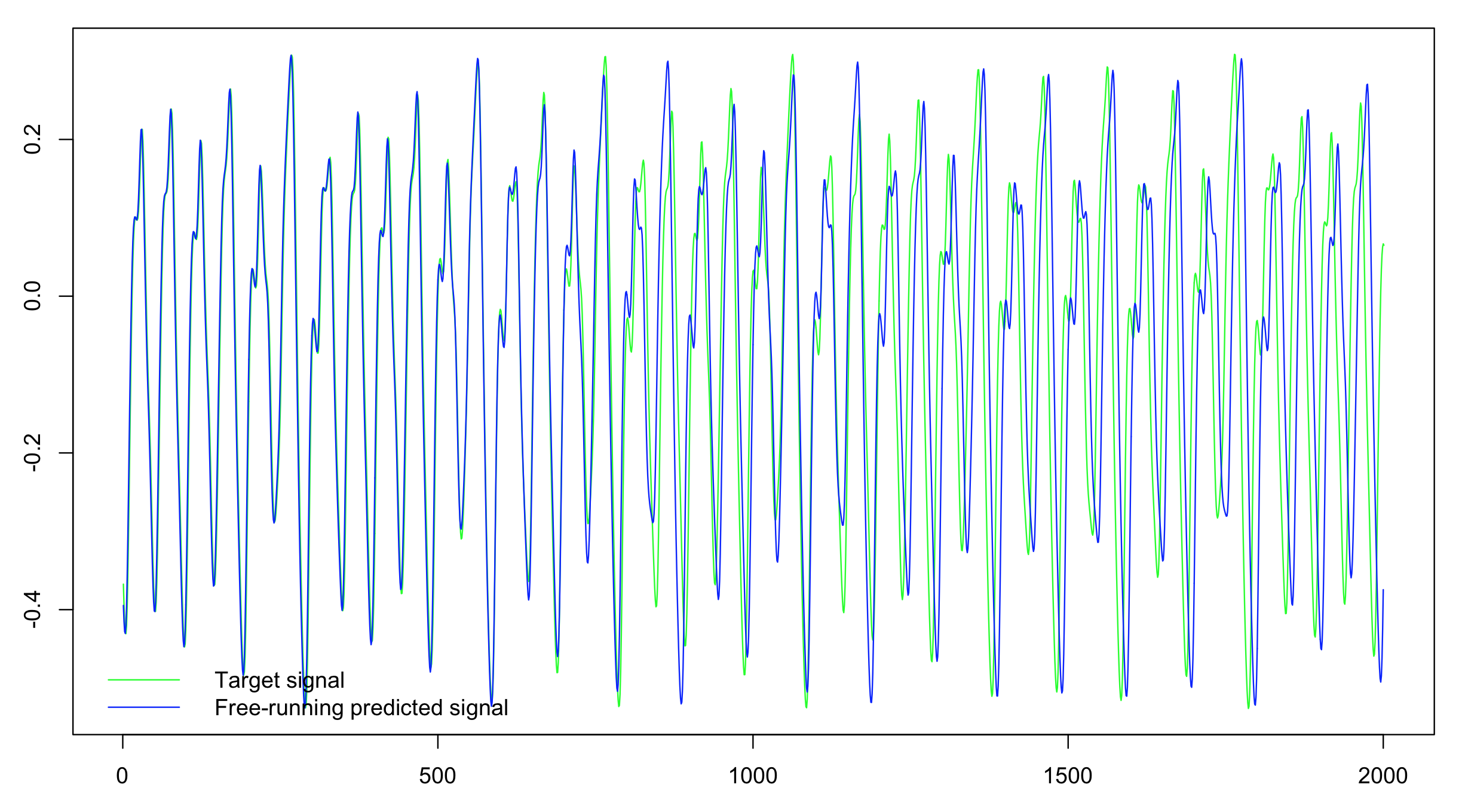}\\
  \caption{Target Signal Tracking for Uniform Distributed Weights}
  \end{center}
\end{figure}

\subsection{Gaussian Distributed Weights}
Two elements are taken into account for the comparison between the Uniform distributed weights and Gaussian distributed weights. We control the shape (range) of the distribution or the variance of the distribution to see which kind of Gaussian distributed weights generate better performance.

\subsubsection{Gaussian Distributed Weights Same Variance}
To control the variance, it means the weights of uniform distribution and weights of Gaussian distribution have the same variance. In this case, the variance is $\sigma^2=1/12$. So the weights follow normal distribution $\mathcal{N}(0, \frac{1}{12})$. 
It gives a better result than the uniform one. The MSE for this case is $9.581 \times 10^{-6}$ while the uniform one is $2.099 \times 10^{-5}$
And from the signal tracking plot Fig. 6., it shows this machine can predict the signal quite well.
\begin{figure}[H]
  \begin{center}
  \includegraphics[width=4in]{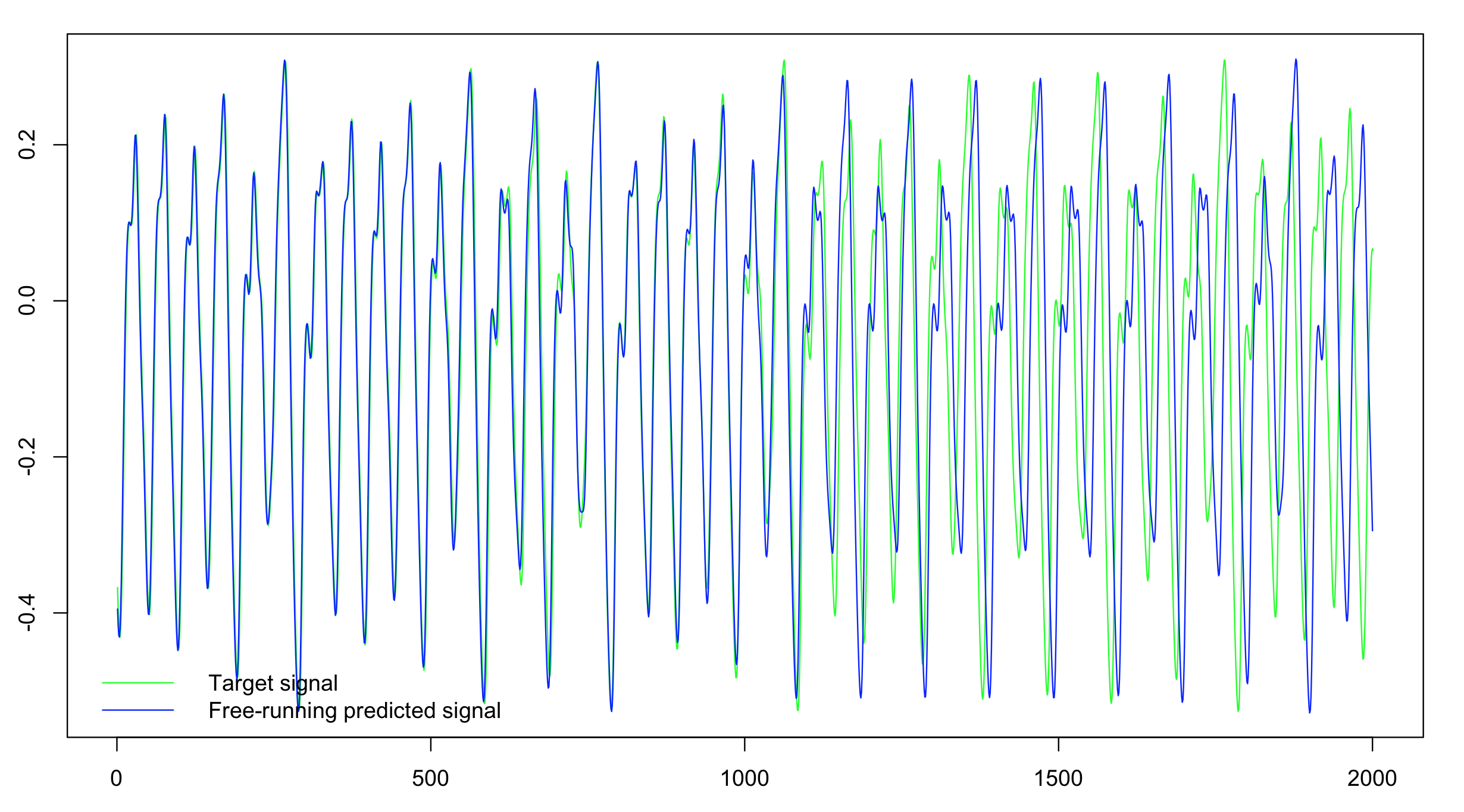}\\
  \caption{Signal Tracking: Gaussian Distributed Weights with Same Variance}
  \end{center}
\end{figure}

\subsubsection{Gaussian Distributed Weights Same Range}
For Gaussian distribution with the control of the range,  we try to keep the weights of uniform distribution and weights of Gaussian distribution in almost the same range $[-0.5,0.5]$. Because for Gaussian Distribution $x\in (-\infty, +\infty)$ is in the open range, we set 99.7\% of the data within the range $[-0.5,0.5]$ , which means points that fall more than 3 standard deviations from the norm are likely outliers. So $3\sigma= 0.5, \sigma^2=1/36$.
\begin{figure}[H]
  \begin{center}
  \includegraphics[width=4in]{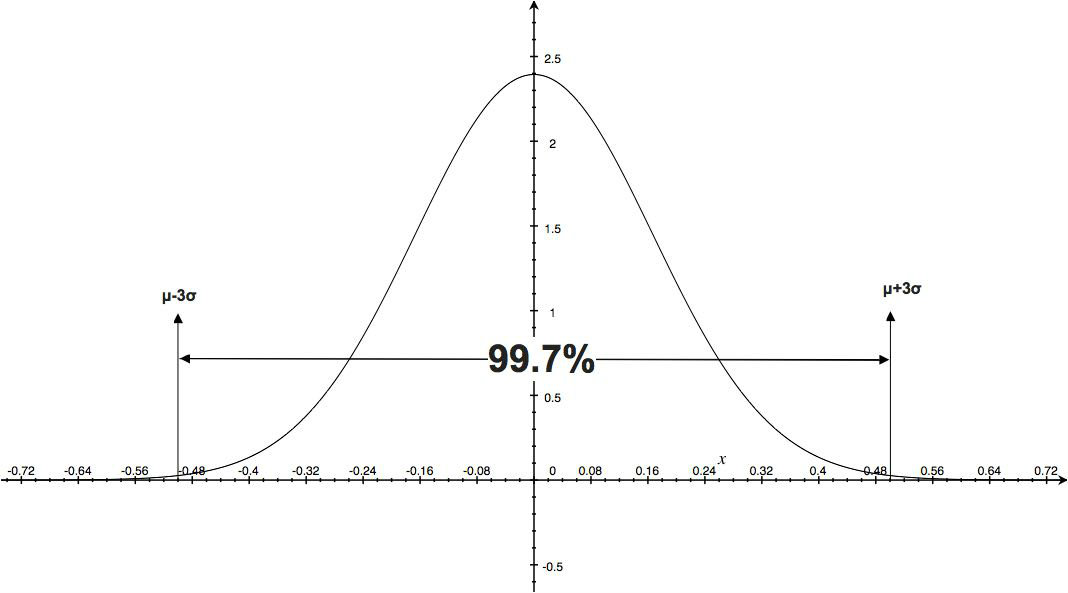}\\
  \caption{ Three-Sigma Rule of Thumb }
  \end{center}
\end{figure}

And we get MSE for Gaussian distributed weights machine $0.0219 $. MSE is much larger than the previous two, showing this kind of weights has poor performance. And from the signal tracking plot Fig. 8., it obviously reflects that the signal can be tracked at the beginning but collapse in the middle part. So this distribution cannot track the signal very well. It may be useful in short-term tracking, but cannot be used in the long run.
\begin{figure}[H]
  \begin{center}
  \includegraphics[width=4in]{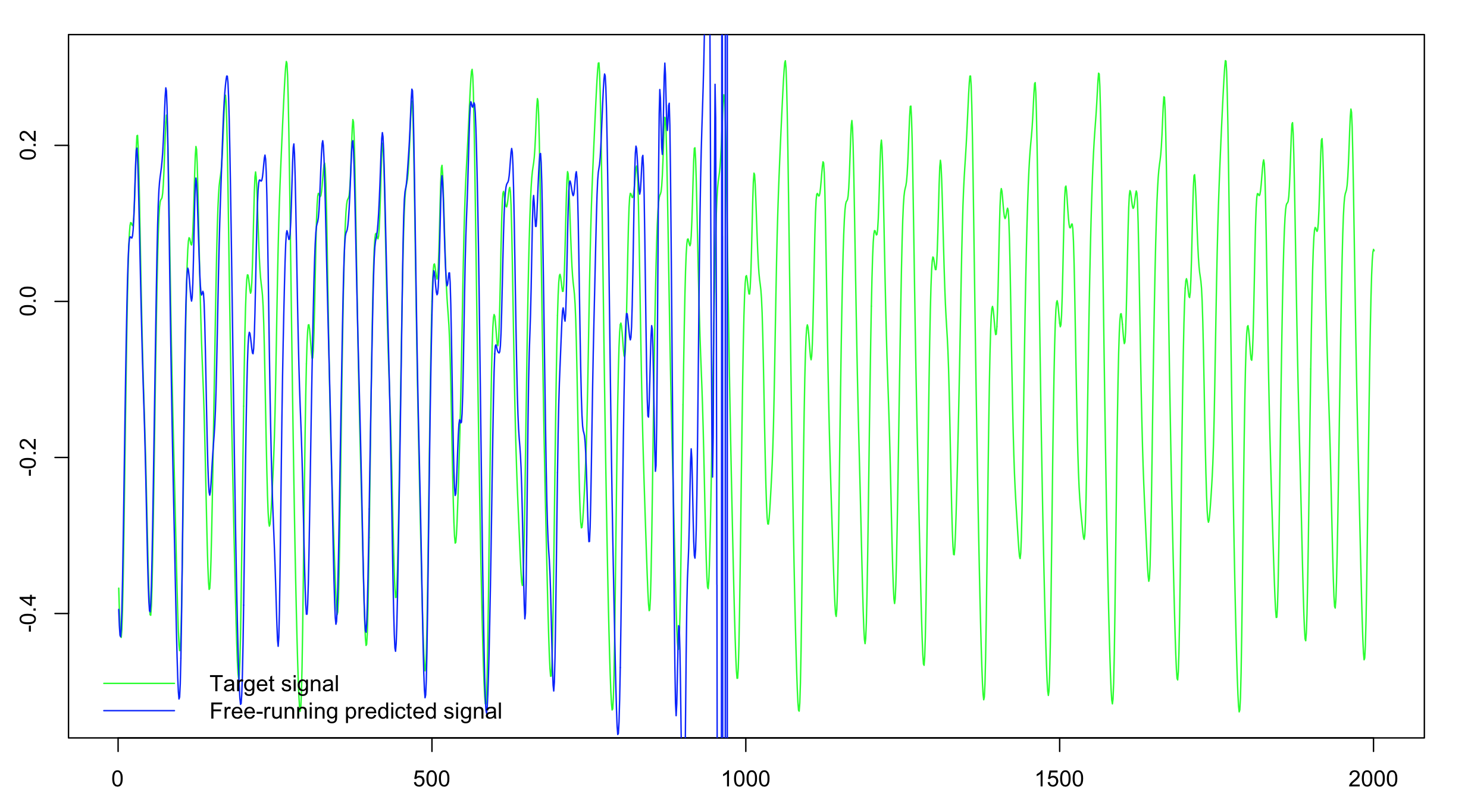}\\
  \caption{Signal Tracking: Gaussian Distributed Weighted with Same Range}
  \end{center}
\end{figure}

\subsection{Arcsine Distributed Weights}
\newcommand{\vw}{\mathbf{w}} 
To explore the distribution of weights with different shapes, Arcsine distribution cannot be ignored as it has an absolutely distinct shape than other distributions.
Technically, a continuous random variable $W$ is said to have an arcsine distribution on $\left[-\frac{1}{2}, +\frac{1}{2}\right]$, if its probability density function is given by (13), it can also be easily established that the CDF for $W$ is given by (14)
\begin{align}
 f_W(\vw)&=\frac{2}{\pi\sqrt{(1-2\vw)(1+2\vw)}}\\ 
  F_W(\vw)&= \frac{2}{\pi} \texttt{arcsine}\Bigg(\sqrt{\frac{2\vw+1}{2}}\Bigg)
\end{align}
In fact, in general, one can define arcsine random variables in $[a, a+l]$, with pdf as (15), and the corresponding CDF for $W$ given by (16)
\begin{align}
     f_W(\vw)&=\frac{1}{\pi\sqrt{(-a+\vw)(a+l-\vw)}}\\
    F_W(\vw) &= \frac{2}{\pi} \texttt{arcsine}\Bigg(\sqrt{\frac{-a+\vw}{l}}\Bigg)
\end{align}
Probability density function of Arcsine distribution is a symmetric U-shaped curve, centered on $(x_{max} - x_{min})/2$, with highest close to the two extrema, and quite flat over the central region. So the shape for Arcsine distribution is quite unique. In this case we get MSE of weights with the distribution that follows arcsine(-0.5, 0.5). MSE here is $ 1.746 \times 10^{-6}$ even smaller than the one from Gaussian distribution with the same variance, showing Arcsine distributed weights have edge over other distributed weights. And from the signal tracking plot Fig. 9., the signal can be tracked super well. So for the time-series data Mackey-Glass (delay=17), weights with Arcsine distribution perform the best.
\begin{figure}[H]
  \begin{center}
  \includegraphics[width=4in]{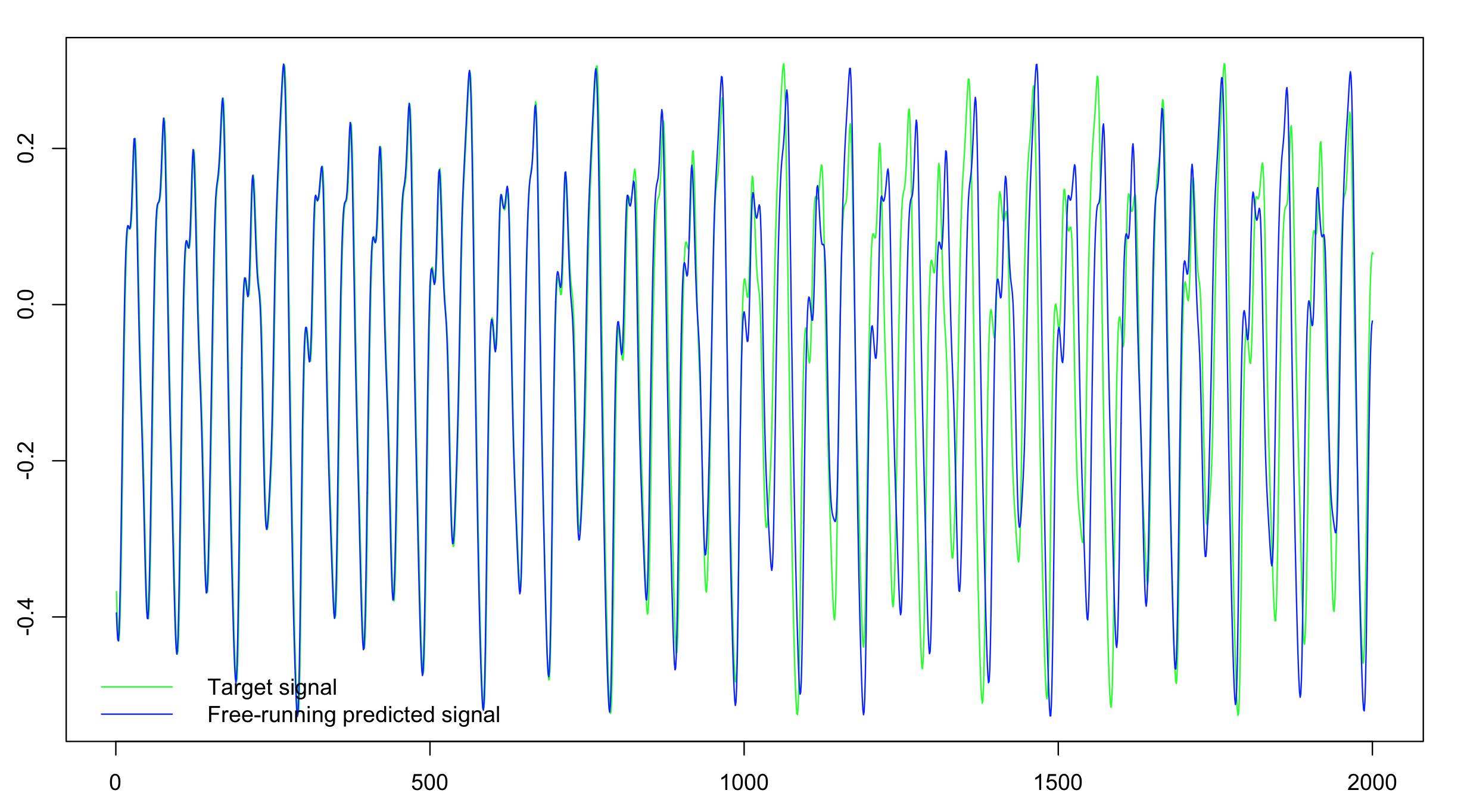}\\
  \caption{Signal Tracking: Arcsine Distributed Weighted}
  \end{center}
\end{figure}

\subsection{Simulation Based on Four Weight Distributions}
For further explore the properties of different distributions of the weights, we generate various time series dataset:
\begin{itemize}
\item Non-stationary time series data without trend
\begin{align}
 X_t = \epsilon_t + 0.81X_{t-1}+0.72\epsilon_{t-1}
\end{align}
\item Non-stationary time series data with trend
\begin{align}
 X_t = \epsilon_t + 0.81X_{t-1}+0.72\epsilon_{t-1} + \frac{t}{1000} + \Big(\frac{t}{1000}\Big)^2 
\end{align}
\item Stationary time series data without trend
\begin{align}
 X_t = \texttt{sin}((1+t)\pi^3)
\end{align}
\item Stationary time series data with trend
\begin{align}
 X_t =\texttt{sin}((1+t)\pi^3) + \frac{t}{1000} 
\end{align}
\end{itemize}
\begin{figure}[H]
  \begin{center}
  \includegraphics[width=6in]{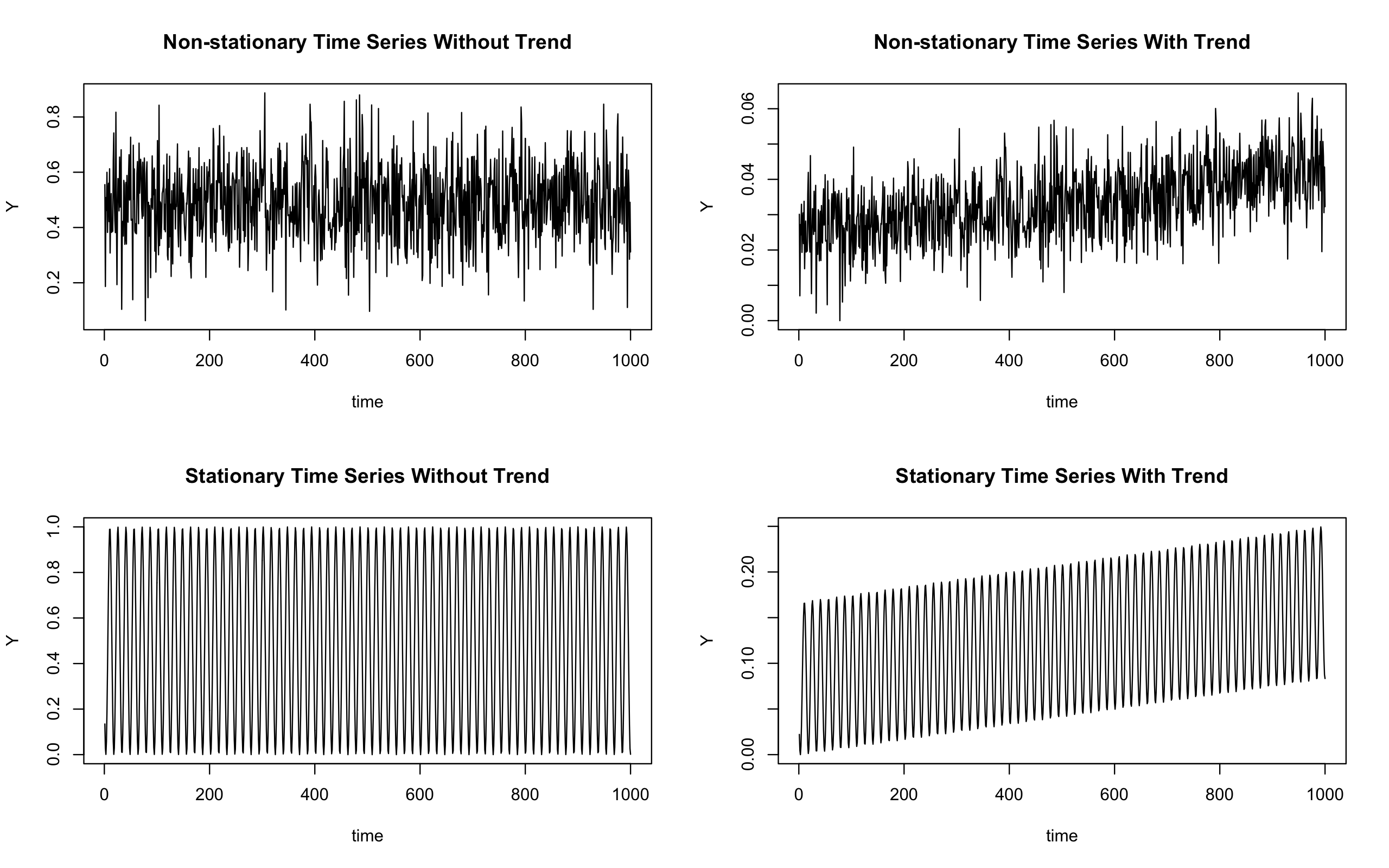}\\
  \caption{Different Time Series Datasets}
  \end{center}
\end{figure}

After intense exploration on four different scenarios we found that for non-stationary time-series data, none of the distribution can handle them and all of them explode. For stationary time-series data without trend, Uniform distributed weight gives the best performance, and other distributed weights generate pretty good prediction as well. For original time-series data and stationary time-series data with trend, Arcsine distributed weight can better track the signal with smaller error. In the table below we denote original Mackey Glass (delay=17) data as MG, stationary time series data without trend as SWT, stationary time series data with trend as ST, Gaussian-A means Gaussian distribution with same variance and Gaussian-B means Gaussian distribution with same range.
\begin{table}[H] \centering 
  \caption{Result of Exploration on Weight Distribution} 
\begin{tabular}{@{\extracolsep{5pt}} ccccc} 
\\[-1.8ex]\hline 
\hline \\[-1.8ex] 
 MSE & Uniform & Gaussian-A & Gaussian-B& Arcsine\\ 
\hline \\[-1.8ex] 
 MG & 2.099$\times 10^{-5}$ & 9.581$\times 10^{-6}$ & 0.0219  & 1.746$\times 10^{-6}$ \\ 
  SWT &   $1.804 \times 10^{-6}$  & $ 5.479 \times 10^{-6}$ & $2.237 \times 10^{-5} $ & 3.100$\times 10^{-5}$ \\ 
  ST & $2.221 \times 10^{-5}$  & $9.030 \times 10^{-6}$ & 0.00439 &  $1.253 \times 10^{-6}$\\ 
\hline \\[-1.8ex]  
\end{tabular} 
\end{table}

From the table above we notice that the shape of the distribution indeed has impact on the accuracy of the tracking. Particularly, the arcsine distribution appears to offer a better result probably due to induce some level of sparsity, which in turn reduces the error.

\section{Tuning Exploration}
As mentioned in second chapter about the environmental setting challenge for Echo State Network, here we will continue and elaborate this topic to different kinds of time series datasets to see how the environmental settings change according to different time series datasets, and how to select the optimal strategy based on the performance from different datasets.

\subsection{Performance}
After exploring different time series datasets with different traits listed above, we discover that Echo State Networks can give decent performance on the majority of stationary dataset without surprise. Additionally, it can also track the information of some non-stationary datasets in short-term. But the machine breaks down in the long run for most non-stationary datasets. Table 2 shows the comparative predictive performance on various time series datasets. To thoroughly explore the performance of ESNs, we also compare ESNs with other common methods for time series prediction such as feed-forward neural networks (NN) and autoregressive integrated moving average model (ARIMA). Based on different dataset for ESN method, we first adjust the datasets for the optimal setting via data preprocessing and parameter turning.
\begin{figure}[H]
  \begin{center}
  \includegraphics[width=2.5in]{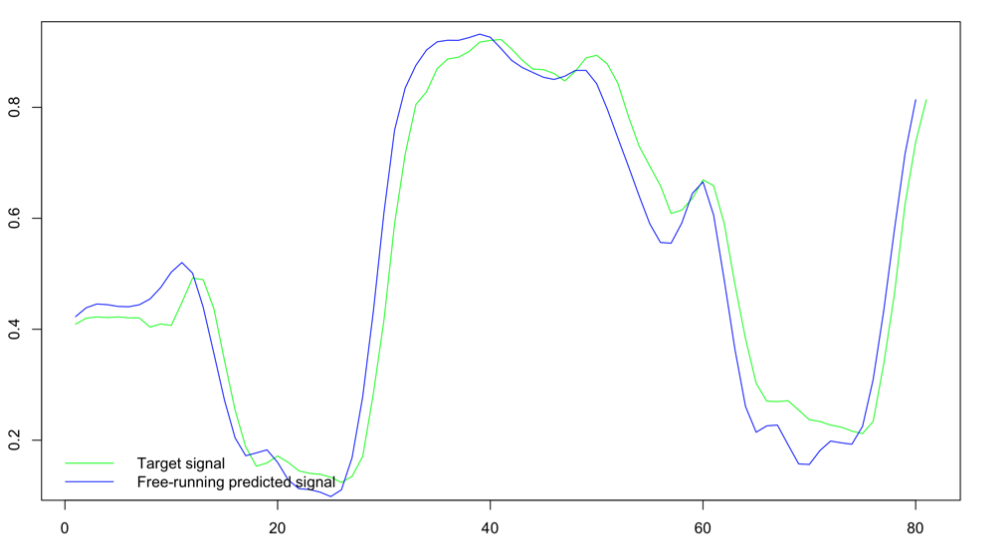}
  \includegraphics[width=2.5in]{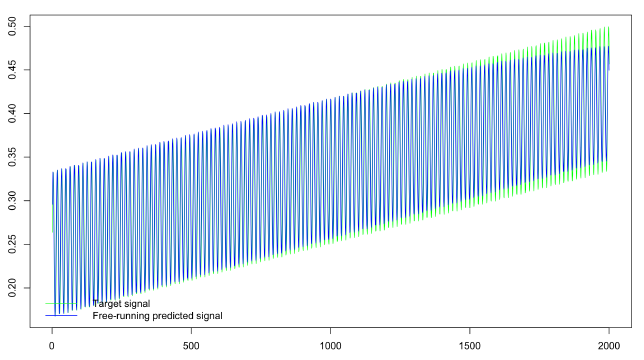}
  \caption{Good Performance of ESN Tracking}
  \end{center}
\end{figure}
\begin{table}[H] \centering 
  \caption{Comparative Predictive Performance on Relative Datasets} 
\begin{tabular}{@{\extracolsep{5pt}} ccccc} 
\\[-1.8ex]\hline 
\hline \\[-1.8ex] 
 Dataset  & Tuning Setting & $MSE_{ESN}$ & $MSE_{NN}$ & $MSE_{ARIMA}$ \\ 
\hline \\[-1.8ex]
 lynx & \makecell{ cubitize\\ dynamic leaking rate\\pertubation on initialization}  & 0.0848 &0.0492 &  0.0615 \\ [0.5ex]
 \hline
  gas  & \makecell{ cubitize\\ dynamic leaking rate\\pertubation on each time step}& 0.0035 & 0.0006 & 0.0008 \\ 
 \hline
  sunspots  & \makecell{ scaling\\ dynamic leaking rate\\pertubation on initialization}& 3.9231 & 0.0103 & 0.0173  \\  
 \hline
  AirPassengers & \makecell{ cubitize\\ dynamic leaking rate\\pertubation on initialization\\pertubation on each time step}& 0.0062 & 0.0415  & 0.1037 \\  
 \hline
 wineind  & \makecell{ cubitize\\ dynamic leaking rate\\pertubation on each time step}& 0.0794& 0.0682 & 0.0458 \\  
 \hline
 taylor  & cubitize & 0.0011  & 0.1760 & 0.0756 \\  
 \hline
 gold  & \makecell{ cubitize\\ dynamic leaking rate\\pertubation on initialization\\pertubation on each time step}&   0.0099 & 0.0078 & 0.096 \\  
 \hline
 woolyrnq  & \makecell{ cubitize\\ dynamic leaking rate}& 0.1672 & 0.0205 &  0.04671 \\  

\hline \\[-1.8ex] 
\end{tabular} 
\end{table} 
For feed-forward neural network and AIRMA method, we haven't extended our work on their environmental setting so the result is not their optimal performance. However, the three methods show the results almost in the same magnitude, which indicates that ESN generates reasonable results as a type of neural network. Furthermore, more efforts are desired for generating seemly performance. Efficient learning environment setting includes:
data preprocessing via cubitization and scaling; perturbation on initialization; perturbation for weights; dynamic leaking rate etc.
In summary, the system collapses when the dataset contains too much noise. The size of the dataset matters and ESN tends to give poor performance on small dataset for lack of information. The machine can track in short term for most dataset, while it cannot handle long-term non-stationary time series dataset. Furthermore, based on different dataset, different methods would be employed. Some need more randomness (perturbation effect, modification on
initialization etc.), while some need to lower the variation (ensemble building, dynamic leaking rate etc.)

\section{Ensemble of Echo State Networks}
Based on the above exploration on the performance from different datasets, we can see that even under the ideal scenarios, the randomness still yields high variance on estimators. The challenge we encounter drive us to look for the potential for Ensembles such as Bagging \cite{breiman_1996} and Random Forest \cite{breiman_last_rice_2000} as an approach to reduce variance and thus generate better results. Bagging, Boosting, Random Forest have in common their inherent capacity to achieve Bias-Variance trade-off and hence have a reduction in prediction error. Especially in big data situations, selecting a single model cannot provide the optimal prediction. Except random forest and boosting in some degree, ensemble methods do not give variable selection or variable importance measure. The reason for this is that the motivation of ensemble learning is predictive optimality instead of selection of a single model.

The basic idea of Ensemble Echo State Networks is that based on the given data $\mathcal{D} = \{(\mathbf{x_i}^T,\mathbf{y_i})^T, i=1,...,n\}$, we want to find out a true but unknown function $f$ that models the functional relationship between variable $X$ and $Y$. Here the function $f$ represents Echo State Networks $f^{(esn)}$. Assume we are given $M$ different estimators of $f^{(esn)}$:
\begin{align}
\mathcal{E} = \{\hat{f}_1, \hat{f}_2, ... , \hat{f}_M   \}
\end{align}
and assume that each estimator $\hat{f}_m(\cdot)$ has a corresponding relative importance measured by weights $\alpha_m$ for $m=1,...,M$. So the goal is to find an estimator of $f$ with the smallest prediction error. That is to say, ensemble learning is to combine all the available decent candidate estimators, and $  \hat{f}(\cdot)$ is a weighted average of base learners:
\begin{align}
\hat{f}^{(ENS)}(\cdot) = \sum_{m=1}^M \alpha_m \hat{f}_m(\cdot)
\end{align}
Each of the estimator  $\hat{f}_m(\cdot)$ is referred to as a base learner. If the base estimators  $\hat{f}_m(\cdot), m =1,...,M$ are strongly correlated, the performance of the ensemble will probably decrease due to the redundant base learner results. Hence, it is significant to select the base learners and decide the weights based on their performance. Another point worthy of note is that the ensemble estimator $\hat{f}^{(ENS)}(\cdot)$ ought to be at least as good as the best of the base learners:
\begin{align}
\overline{err}( \hat{f}^{(ENS)}(\cdot))\leq \underset{m=1,...,M}{argmin}\{\overline{err}(\hat{f}_m(\cdot) )\}
\end{align}
Therefore, how to build each $\hat{f}_m(\cdot)$ and how to find $\hat{\alpha_m}$ are the key points for good ensembles. There are two kinds of base learners: homogeneous base learners and heterogeneous base learners depending on if all base learners come from the same function class. For heterogeneous base learners, one can combine different kinds of base learners such as SVM classifiers and trees. For homogeneous base learners, all base learners are from the same class such as all from linear models with each has a different subset of variables trained on different random subsets of the same data, or all from trees built on random subsets of the same data as it is the case in tree bagging and random forest.
Regardless of how the base learners are obtained, the resulting aggregation is equation (26).
Here we use homogeneous base learners to build ensembles because the base learners functions are all from Echo State Networks. Here we mainly explode two scenarios for building ensemble base learners: Perturbation of Weights and Bootstrap Aggregating.

\subsection{Perturbation of Weights}
First scenario is perturbation of the input weights $\mathbf{W}^{in}$ and the weights in hidden layer $\mathbf{W}$. If $\mathbf{W}^{in}$ and $\mathbf{W}$ change at each step, then the corresponding $\hat{f}_m(\cdot)$ changes. So many copies of $\hat{f}_m(\cdot)$ can be generated by changing weights $W^{in}$ and $W$. The step for this scenario:
\begin{itemize}
\item Generate $W^{in}_m$ and $W_m$
\item Form $\hat{f}_{m}(\cdot)$ for every perturbation of weights
\item Build ensemble on the above $\hat{f}_m(\cdot)$
$$ \mathcal{E}(W^{in}, W) = \{ \hat{f}_m(\cdot), m=1,2,...,M, W_m^{in}, W_m  \}$$
\end{itemize}

\begin{algorithm}[H]
\caption{Perturbation on Weights}
\begin{algorithmic} 
\FOR{$m \gets 1:M$}
\STATE Generate $W^{in}$ and $W$ with fixed dataset
\STATE Form $\hat{f}_{m}(\cdot)$ based on different $W_m^{in}$ and $W_m$
\ENDFOR
\STATE Build ensemble: $\hat{f}^{ENS}(u^{new}) = \{\hat{f}_m(u^{new}), m=1,2,...,M \}$
\end{algorithmic}
\end{algorithm}

\subsection{Bootstrap Aggregating}
Bootstrap aggregating is one of machine learning ensemble algorithm designed to improve the stability and accuracy in statistical classification and regression. It also reduces variance and helps to avoid overfitting. It is usually used in decision tree methods, but it can be applied with any type of method. Bootstrap aggregating was proposed by Leo Breiman in 1994\cite{breiman_1994}. His motivation was to achieve variance reduction in classification and regression, and the aggregation of many replicates of an estimator is more stable than a single instance of that estimator. 

Given dataset $\mathcal{D} = \{(x_i,y_i), i=1,2,...,n \}$ along with $M$ estimators $\hat{f}_{m}(\cdot), m = 1,2,...,M $ of $f: \mathcal{X} \rightarrow  \mathcal{Y}$, instead of choosing $1$ out of the $M$, we aggregate all the $M$ $\hat{f}_{m}(\cdot)$. To get $\hat{f}_{m}(\cdot)$, we first use bootstrap to create many copies of the data $\mathcal{D}$, and then for each copy we create $\hat{f}_{m}(\cdot)$. The step for this scenario:
\begin{itemize}
\item Uniformly sample the given training dataset $\mathcal{D}$ of size $n$ with replacement
\item Generate M new training datasets $\mathcal{D}_m$ of size $n'$
\item When $n = n' $ for large n, $\mathcal{D}_m$ has the fraction $ (1 - 1/e)$ of the unique examples of $\mathcal{D}$, and duplicates the rest \cite{Aslam}
\item For each new data $\mathcal{D}_m$ create $\hat{f}_{m}(\cdot)$
\item Build ensemble on the above $\hat{f}_{m}(\cdot)$
\end{itemize}

\begin{algorithm}[H]
\caption{Bootstrap Aggregating (Time Series Dataset)}
\begin{algorithmic} 
\STATE Initialize $m:=0$
\REPEAT
\STATE $ m := m+1$
\STATE Form a bootstrap sample $\mathcal{D}_{m} = \{(x_1^{(m)}, y_1^{(m)}),...,(x_n^{(m)}, y_n^{(m)})\}$ by randomly sampling with replacement n times from $\mathcal{D} = \{(x_1, y_1),...,(x_n, y_n)\}$
\STATE  Arrange the data in each new dataset $\mathcal{D}_{m}$ in time order
\STATE  Construct the bootstrapped estimator $\hat{f}_{m}(\cdot)$ based on $\mathcal{D}_{m}$
\UNTIL{$m=M$}
\STATE Compute the bagged estimator $\hat{f}^{ENS}(u^{new}) = \frac{1}{M}\sum_{m=1}^M\hat{f}_{m}(u^{new})$
\end{algorithmic}
\end{algorithm}
The ensemble estimator $\hat{f}^{ESN}(\cdot)$ is driven by the hyper-parameter $M$, the number of base learners (base echo state networks) in the ensemble. A good way to determine the optimal value of $M$ for a given problem is tuning using cross validation
\begin{align}
M^{(opt)} = \underset{M\in grid(M) }{\texttt{argmin}}\{\texttt{CV} ( \hat{f}^{ESN} )\}
\end{align}
As M increases, $\texttt{Variance}(\hat{f}^{ESN})$ decreases without any increase in $\texttt{Bias}(\hat{f}^{ESN})$ which leads to bias-variance trade-off and optimal predictive performance as a result.

\subsection{Simulations}
For the eight time series datasets we explored in the fourth chapter above, we extend the exploration and build Ensembles of Echo State Networks for these datasets with the approaches of randomization on weights. We found that the Ensembles consistently outperform the base learners. The mean square error of base learner is denoted by $\bar{\mathcal{E}}_{s}$ and the mean square error of ensemble estimator is $\bar{\mathcal{E}}_{ens}$ in the table below.

\begin{table}[H] \centering 
  \caption{MSE Comparison and Error Reduction} 
\begin{tabular}{@{\extracolsep{5pt}} cccc} 
\\[-1.8ex]\hline 
\hline \\[-1.8ex] 
 Dataset & $\bar{\mathcal{E}}_{s}$ & $\bar{\mathcal{E}}_{ens}$ & Error Reduction  \\ 
\hline \\[-1.8ex] 
 lynx & 0.0848  &     0.0661    &  22.05\%   \\
  gas &  0.0035   &    0.0016     &   54.28\%  \\ 
  sunspots &  3.9231  & 0.5120  & 86.95\%       \\  
  AirPassengers & 0.0062   &   0.0058     &  6.45\% \\  
 wineind &   0.0794     &   0.0635    &  20.03\% \\  
 taylor &   0.0011       &  0.0005 &     54.55\%\\  
 gold &  0.0099   &    0.0057     & 42.42\%\\  
 woolyrnq & 0.1672    &     0.1373   &    17.88\% \\ 
\hline \\[-1.8ex] 
\end{tabular} 
\end{table}

The errors drop significantly with the ensemble approach under the optimal base learner scenario we achieved in the previous chapter. As the input weights and the weights in the hidden layer are selected at random, the model is going to perform differently given diverse weights. The prediction result for single ESN inevitably contains uncertainty. Therefore, ensemble ESN with the reasonable number of qualified members can extract the inherent feature through uncertainty and also give credit to relative weak learners. Here we regard each base learner equally weighted.

For ensemble with bootstrap method, the pivot idea is to draw sample from training set (subsets of training dataset). So the dataset itself could not be too small. Here we use large dataset Mackey-Glass (delay=17) time series data from Chapter 3, containing 10000 value. We get:
\begin{align*}
\bar{\mathcal{E}}_{s} &= 2.099\times10^{-5}\\ 
\bar{\mathcal{E}}_{ens} &= 7.865\times10^{-6}\\
 \Delta\bar{\mathcal{E}}&= \left|\frac{7.865\times10^{-6} - 2.099\times10^{-5}}{2.099\times10^{-5}}\right| = 62.53\%
\end{align*}
The result illustrates that the Ensemble of Echo State Networks can improve the performance and reduce the variation caused by randomness if the base learners provide reasonable result. The stability of the model is enhanced by ensemble technique.

\section{Conclusion and Future Work}
Echo state networks are compelling recurrent neural networks with the sparse random connections in the hidden layer reflecting the previous states. Nevertheless, the process of obtaining a decent ESN for particular datasets is pretty tough due to its instability. ESN is notoriously difficult to track data because of the following reasons:
\begin{itemize}
\item The performance of the base learner is quite unsettling, which demands tremendous efforts for tweaking the parameter in the initial setting stage.
\item The system would collapse when the dataset contains much noise.
\item ESN can only track temporarily and it unable to forecast in the long run.
\end{itemize}
In order to conquer this challenging we create, develop and implement a family of predictably optimal robust and stable ensemble of Echo State Networks by regularizing the training and perturbing the input. In addition, several distributions of weights with significant features have been studied to see if the shape of the distribution has the effect for reducing the error. We notice that the shape of the distribution indeed has impact on the accuracy of the tracking. Particularly, the arcsine distribution appears to generate better results due to the level of sparsity, which in turn reduces the error.

During the exploration on the environmental setting we conclude that the efficient learning environment setting includes: data preprocessing via cubitization and scaling; perturbation on initialization; perturbation for weights; dynamic leaking rate. Furthermore, ESN can track in short term for most cases, but it collapses in the long run. Short-term tracking with large size reservoir enables ESN to perform strikingly with superior prediction. 

Based on this scenario, we go a further step to aggregate many of ESNs into an ensemble to lower the variance and stabilize the system. We create the ensemble of Echo State Networks with two scenarios: (i) Adaptation on structured pattern matrices (ii) Stochastic replications and bootstrapping of input data. Ensembles of ESN turn out to give pretty good performance. One thing worthy of note is that the fundamental base learner should be at least give reasonable result so that the ensemble can work and achieve magnificent prediction. In the most basic ensembles, all the base learners are equally weighted, the same as the work in this paper. However, it is not necessary the best way. Finding optimal weights for ensemble model is actually one of the most common challenges. Future work can explore this and also more work needs to be done to build stable base ESNs, so that ensemble model can do well.

\nocite{*}


\nolinenumbers

\bibliography{library}

\bibliographystyle{abbrv}

\begin{appendices}
\section{Datasets}
Totally we study 8 different time series datasets from \texttt{R package library(forecast)} and \texttt{library(datasets)}.
\begin{itemize}
\item \texttt{AirPassengers\{datasets\}}: The classic Box \& Jenkins airline data. Monthly totals of international airline passengers from 1949 to 1960.
\item \texttt{taylor\{forecast\}}: Half-hourly electricity demand in England and Wales from Monday 5 June 2000 to Sunday 27 August 2000. Discussed in \cite{taylor_2003}, and kindly provided by James W Taylor. 
\item \texttt{gold\{forecast\}}: Daily morning gold prices in US dollars from 1 January 1985 to 31 March 1989. \cite{portrait_2010} 
\item \texttt{lynx\{datasets\}}: Annual numbers of lynx trappings from 1821 to 1934 in Canada. Taken from \cite{brockwell_davis_1991}, this appears to be the series considered by \cite{campbell_walker_1977}.
\item \texttt{gas\{forecast\}}: Australian monthly gas production from 1956 to 1995 from Australian Bureau of Statistics.
\item \texttt{sunspots\{datasets\}}: Monthly mean relative sunspot numbers from 1749 to 1983. Collected at Swiss Federal Observatory, Zürich until 1960, then Tokyo Astronomical Observatory. \cite{schmee_1987}
\item \texttt{wineind\{forecast\}}: Australian total wine sales by wine makers in bottles from January 1980 to August 1994. \cite{portrait_2010} 
\item \texttt{woolyrnq\{forecast\}}: Quarterly production of woolen yarn in Australia from March 1965 to September 1994. \cite{portrait_2010} 
\end{itemize}
\begin{figure}[H]
  \begin{center}
    \includegraphics[width=4in]{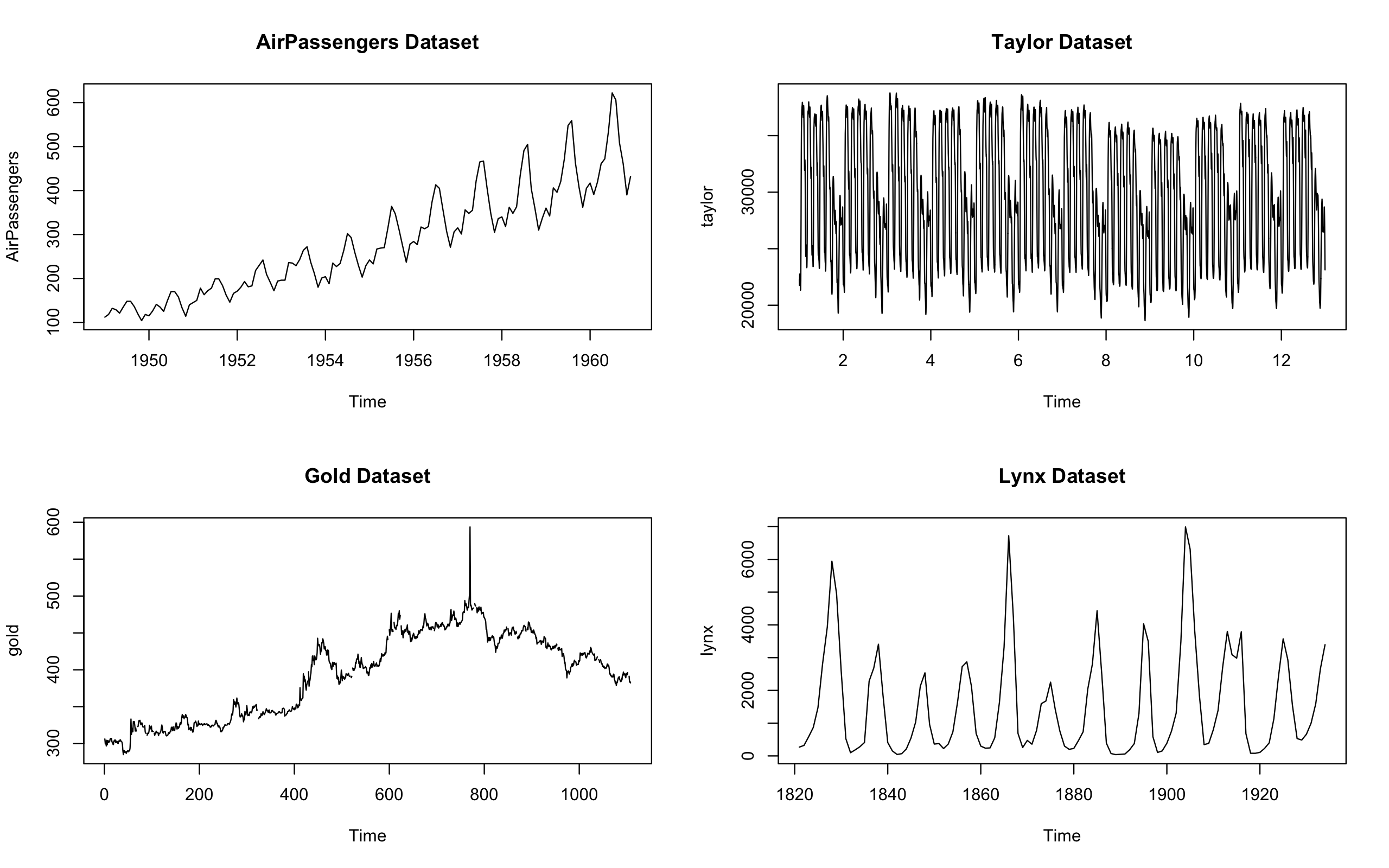}\\
  \includegraphics[width=4in]{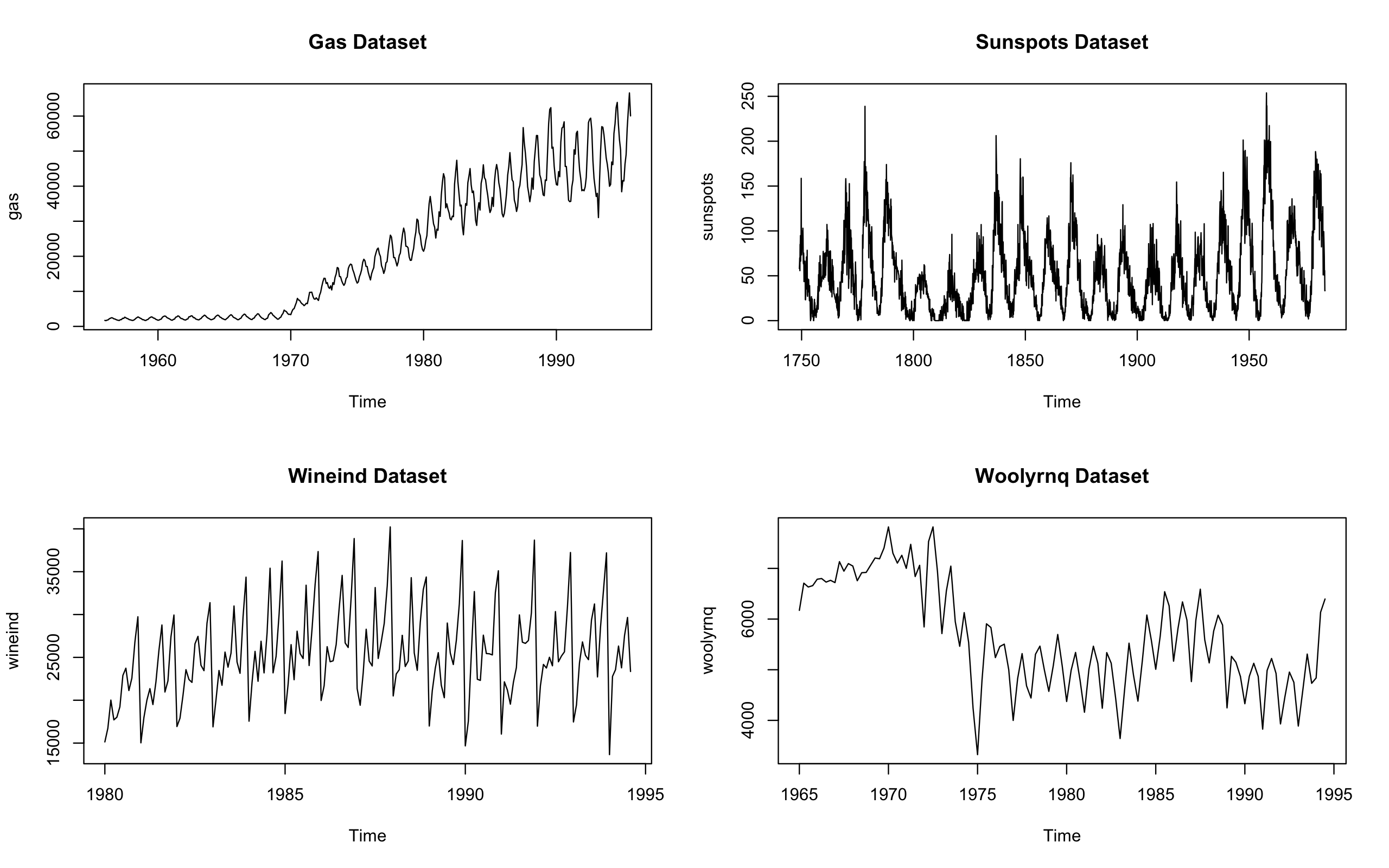}\\
  \caption{Different Time Series Datasets}
  \end{center}
\end{figure}
\end{appendices}

\end{document}